\documentclass[10pt,twocolumn,letterpaper]{article}

\usepackage{btas}
\usepackage{times}
\usepackage{epsfig}
\usepackage{graphicx}
\usepackage{amsmath}
\usepackage{amssymb}

\usepackage{url}

\usepackage{breakurl}
\usepackage{multirow}
\usepackage{adjustbox}

\usepackage[pagebackref=true,breaklinks=true,letterpaper=true,colorlinks,bookmarks=false]{hyperref}

\btasfinalcopy 

\ifbtasfinal\fi
\begin{document}

\title{Multi-task Learning For Detecting and Segmenting Manipulated Facial Images and Videos}

\author{Huy H. Nguyen$^{\star}$, Fuming Fang$^{\dagger}$, Junichi Yamagishi$^{\star\dagger\ddagger}$, and Isao Echizen$^{\star\dagger}$\\
$^{\star}$SOKENDAI (The Graduate University for Advanced Studies), Kanagawa, Japan\\
$^{\dagger}$National Institute of Informatics, Tokyo, Japan\\
$^{\ddagger}$The University of Edinburgh, Edinburgh, UK\\
\small{Email: \{nhhuy, fang, jyamagis, iechizen\}@nii.ac.jp}}

\maketitle
\thispagestyle{empty}

\maketitle
\thispagestyle{empty}

\begin{abstract}
Detecting manipulated images and videos is an important topic in digital media forensics. Most detection methods use binary classification to determine the probability of a query being manipulated. Another important topic is locating manipulated regions (i.e., performing segmentation), which are mostly created by three commonly used attacks: removal, copy-move, and splicing. We have designed a convolutional neural network that uses the multi-task learning approach to simultaneously detect manipulated images and videos and locate the manipulated regions for each query. Information gained by performing one task is shared with the other task and thereby enhance the performance of both tasks. A semi-supervised learning approach is used to improve the network's generability. The network includes an encoder and a Y-shaped decoder. Activation of the encoded features is used for the binary classification. The output of one branch of the decoder is used for segmenting the manipulated regions while that of the other branch is used for reconstructing the input, which helps improve overall performance. Experiments using the FaceForensics and FaceForensics++ databases demonstrated the network’s effectiveness against facial reenactment attacks and face swapping attacks as well as its ability to deal with the mismatch condition for previously seen attacks. Moreover, fine-tuning using just a small amount of data enables the network to deal with unseen attacks.
\end{abstract}

\section{Introduction}
A major concern in digital image forensics is the deepfake phenomenon~\cite{deepfake}, a worrisome example of the societal threat posed by computer-generated spoofing videos. Anyone who shares video clips or pictures of him or herself on the Internet may become a victim of a spoof-video attack. Several available methods can be used to translate head and facial movements in real time~\cite{thies2016face2face, kim2018deep} or create videos from photographs~\cite{averbuch2017bringing, chung2017you}. Moreover, thanks to advances in speech synthesis and voice conversion~\cite{lorenzo2018voice}, an attacker can also clone a person's voice (only a few minutes of speech are needed) and synchronize it with the visual component to create an audiovisual spoof~\cite{suwajanakorn2017synthesizing, chung2017you}. These methods may become widely available in the near future, enabling anyone to produce deepfake material.

Several countermeasures have been proposed for the visual domain. Most of them were evaluated using only one or a few databases, including the CGvsPhoto database~\cite{rahmouni2017distinguishing}, the Deepfakes databases~\cite{afchar2018mesonet, korshunov2018deepfakes, li2018ictu}, and the FaceForensics/FaceForensics++ databases~\cite{rossler2018faceforensics, rossler2019faceforensics++}. Cozzolino et al. addressed the transferability problem of several state-of-the-art spoofing detectors~\cite{cozzolino2018forensictransfer} and developed an autoencoder-like architecture that supports generalization and can be easily adapted to a new domain with simple fine-tuning.

\begin{figure}[t!]
\begin{center}
\includegraphics[width=83mm]{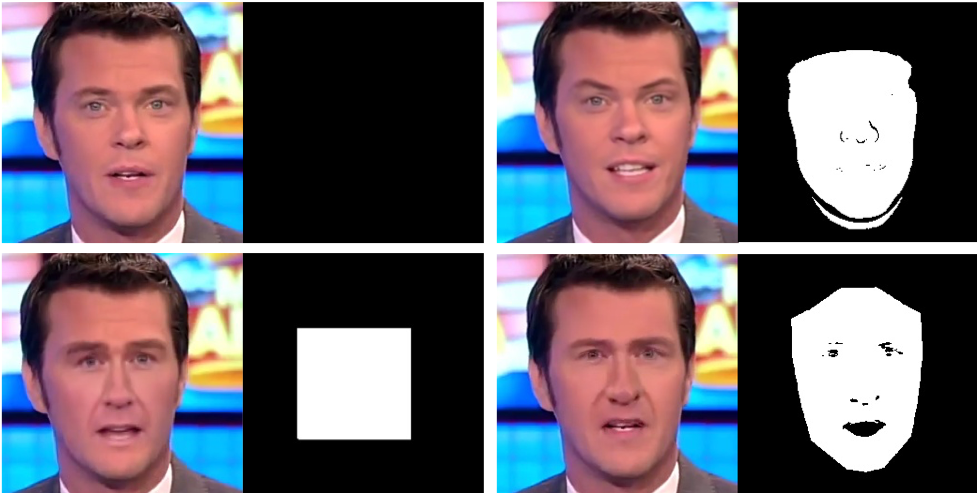}
\caption{Original video frame (top left), video frame modified using Face2Face method~\cite{thies2016face2face} (top right, smooth mask almost completely covers the skin area), using Deepfakes method~\cite{deepfake} (bottom left, rectangular mask), and using FaceSwap method~\cite{rossler2019faceforensics++} (bottom right, polygon-like mask).}
\label{figure:samples}
\end{center}
\end{figure}

Another major concern in digital image forensics is locating manipulated regions. The shapes of the segmentation masks for manipulated facial images and videos could reveal hints about the type of manipulation used, as illustrated in Figure~\ref{figure:samples}. Most existing forensic segmentation methods focus on three commonly used means of tampering: removal, copy-move, and splicing~\cite{bappy2017exploiting, zhou2018learning, bappy2019hybrid}. As in other image segmentation tasks, these methods need to process full-scale images. Rahmouni et al.~\cite{rahmouni2017distinguishing} used a sliding window to deal with high-resolution images, as subsequently used by Nguyen et al.~\cite{nguyen2018modular} and Rossler et al.~\cite{rossler2018faceforensics}. This sliding window approach effectively segments manipulated regions in spoofed images~\cite{rossler2018faceforensics} created using the Face2Face method~\cite{thies2016face2face}. However, these methods need to score many overlapped windows by using a spoofing detection method, which takes a lot of computation power.

We have developed a multi-task learning approach for simultaneously performing classification and segmentation of manipulated facial images. Our autoencoder comprises an encoder and a Y-shaped decoder and is trained in a semi-supervised manner. The activation of the encoded features is used for classification. The output of one branch of the decoder is used for segmentation, and the output of the other branch is used to reconstruct the input data. The information gained from these tasks (classification, segmentation, and reconstruction) is shared among them, thereby improving the overall performance of the network.

\section{Related Work}
\subsection{Generating Manipulated Videos}
Creating a photo-realistic digital actor is a dream of many people working in computer graphics. One initial success is the Digital Emily Project~\cite{alexander2010digital}, in which sophisticated devices were used to capture the appearance of an actress and her motions to synthesize a digital version of her. At that time, this ability was unavailable to attackers, so it was impossible to create a digital version of a victim. This changed in 2016 when Thies et al. demonstrated facial reenactment in real time~\cite{thies2016face2face}. Subsequent work led to the ability to translate head poses~\cite{kim2018deep} with simple requirements that are met by any normal person. The Xpression mobile app\footnote{https://xpression.jp/} providing the same function was subsequently released. Instead of using RGB videos as was done in previous work~\cite{thies2016face2face, kim2018deep}, Averbuch et al. and Chung et al. used ID-type photos~\cite{averbuch2017bringing, chung2017you}, which are easily obtained on social networks. Combining this capability with speech synthesis or voice conversion techniques~\cite{lorenzo2018voice}, attackers are now able to make spoof videos with voices~\cite{suwajanakorn2017synthesizing, chung2017you}, which are more convincingly authentic.

\subsection{Detecting Manipulated Images and Videos}
Several countermeasures have been introduced for detecting manipulated videos. A typical approach is to treat a video as a sequence of image frames and work on the images as input. The noise-based method proposed by Fridrich and Kodovsky~\cite{fridrich2012rich} is considered one of the best handcrafted detectors. Its improved version using a convolutional neural network (CNN)~\cite{cozzolino2017recasting} demonstrated the effectiveness of using automatic feature extraction for detection. Among deep learning approaches to detection, fine-tuning and transfer learning take advantage of high-performing pre-trained models~\cite{raghavendra2017transferable, rossler2018faceforensics}. Using part of a pre-trained CNN as the feature extractor is an effective way to improve the performance of a CNN~\cite{nguyen2018modular, nguyen2018capsule}. Other approaches to detection include using a constrained convolutional layer~\cite{bayar2016deep}, using a statistical pooling layer~\cite{rahmouni2017distinguishing}, using a two-stream network~\cite{zhou2017two}, using a lightweight CNN network~\cite{afchar2018mesonet}, and using two cascaded convolutional layers at the bottom of a CNN~\cite{quan2018distinguishing}. Cozzolino et al. created a benchmark for determining the transferability of state-of-the-art detectors for use in detecting unseen attacks~\cite{cozzolino2018forensictransfer}. They also proposed an autoencoder-like architecture with which adaptation ability was greatly increased. Li et al. proposed using a temporal approach and developed a network for detecting eye blinking, which is not well reproduced in fake videos~\cite{li2018ictu}. Our proposed method, besides performing classification, provides segmentation maps of manipulated areas. This additional information could be used as a reference for judging the authenticity of images and videos, especially when the classification task fails to detect spoofed inputs.

\subsection{Locating Manipulated Regions in Images}
There are two commonly used approaches to locating manipulated regions in images: segmenting the entire input image and repeatedly performing binary classification using a sliding window. The segmentation approach is commonly used to detect removal, copy-move, and splicing attacks~\cite{bappy2017exploiting, bappy2019hybrid}. Semantic segmentation methods~\cite{long2015fully, badrinarayanan2017segnet} can also be used for forgery segmentation~\cite{bappy2019hybrid}. A slightly different segmentation approach is to return the boxes that represent the boundaries of the manipulated regions instead of returning segmentation masks~\cite{zhou2018learning}. The sliding window approach is used more for detecting spoofing regions generated by a computer to create spoof images or videos from bona fide ones~\cite{rahmouni2017distinguishing, nguyen2018modular, rossler2018faceforensics}. In this approach, binary classifiers for classifying images as spoof or bona fide are called at each position of the sliding window. The stride of the sliding window may equal the length of the window (non-overlapped)~\cite{rahmouni2017distinguishing} or be less than the length (overlapped)~\cite{nguyen2018modular, rossler2018faceforensics}). Our proposed method takes the first approach but with one major difference: only the facial areas are considered instead of the entire image. This overcomes the computation expense problem when dealing with large inputs.

\begin{figure*}[t!]
\begin{center}
\includegraphics[width=115mm]{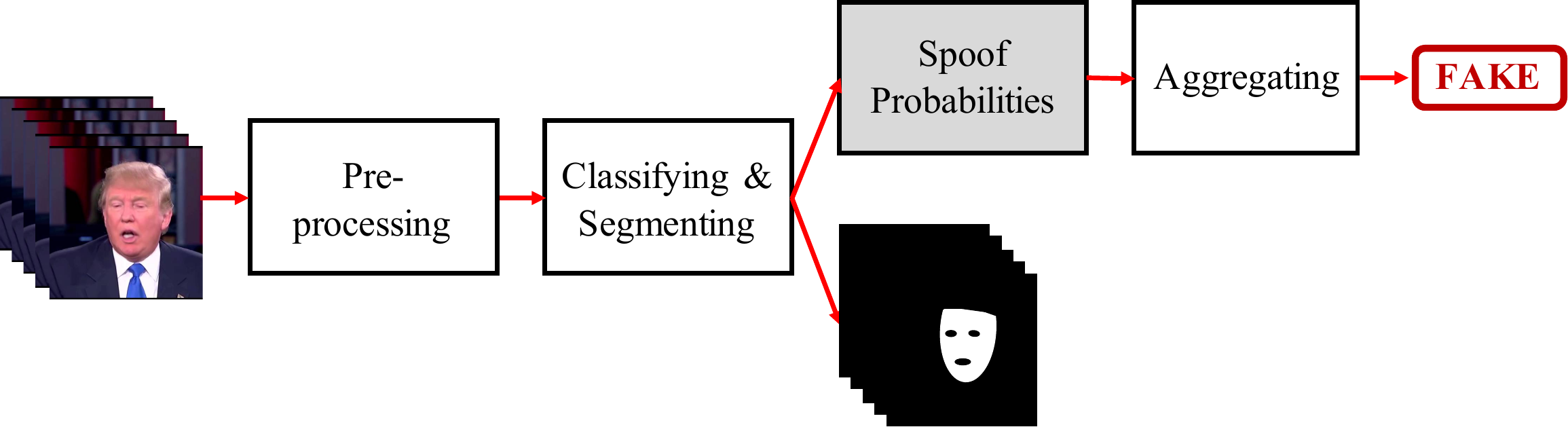}
\caption{Overview of proposed network.}
\label{figure:overview}
\end{center}
\end{figure*}

\section{Proposed Method}
\subsection{Overview}

Unlike other single-target methods~\cite{nguyen2018capsule, cozzolino2018forensictransfer, bappy2019hybrid}, our proposed method outputs both the probability of an input being spoofed and segmentation maps of the manipulated regions in each frame of the input, as diagrammed in Figure~\ref{figure:overview}. Video inputs are treated as a set of frames. We focused on facial images in this work, so the face areas are extracted in the pre-processing phase. In theory, the proposed method can deal with various sizes of input images. However, to maintain simplicity in training, we resize cropped images to $256 \times 256$ pixels before feeding them into the autoencoder. The autoencoder outputs the reconstructed version of the input image (which is used only in training), the probability of the input image having been spoofed, and the segmentation map corresponding to this input image. For video inputs, we average the probabilities of all frames before drawing a conclusion on the probability of the input being real or fake.

\begin{figure}[th!]
\begin{center}
\includegraphics[width=82mm]{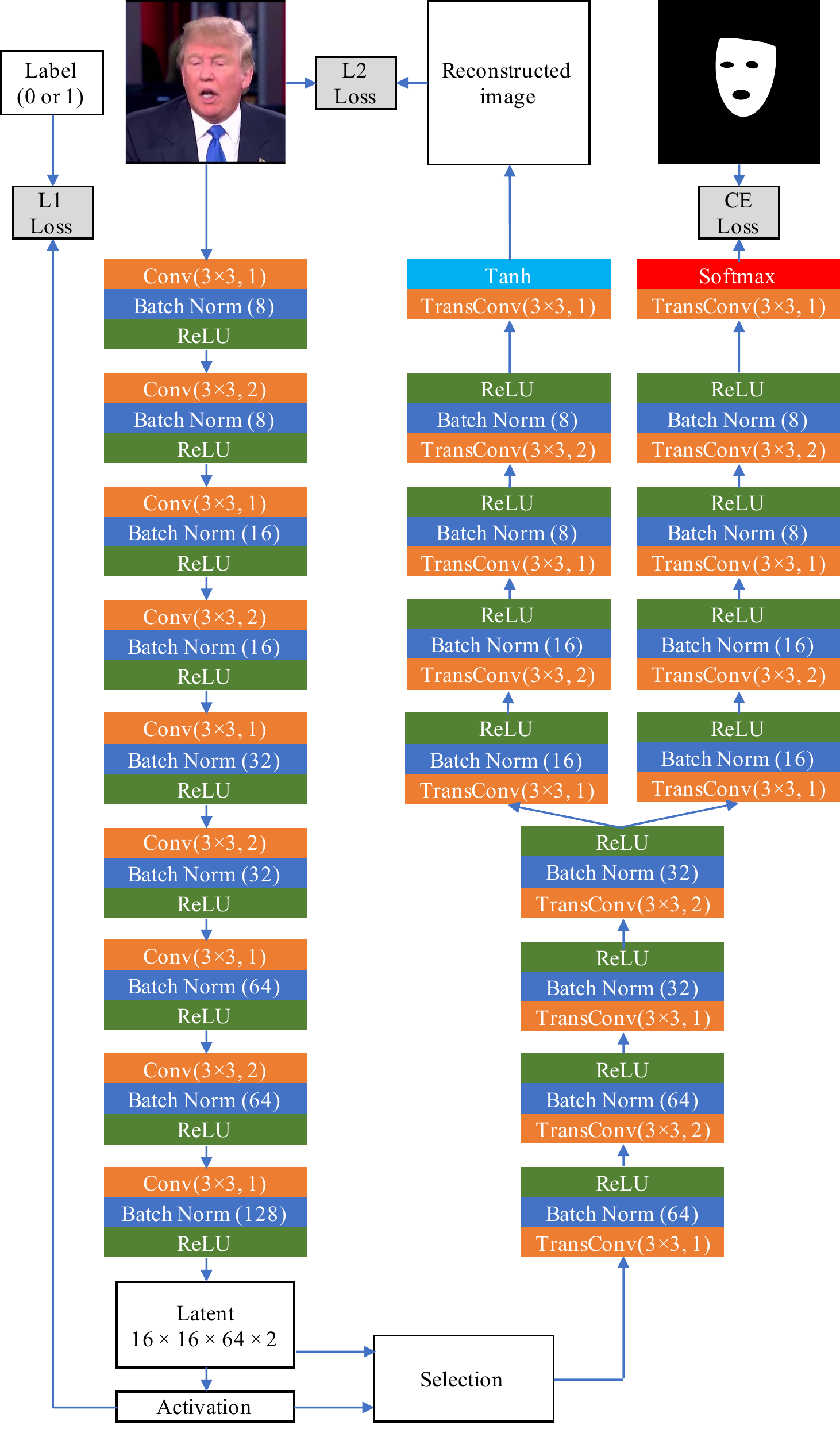}
\caption{Proposed autoencoder with Y-shaped decoder for detecting and segmenting manipulated facial images.}
\label{figure:network}
\end{center}
\end{figure}

\subsection{Y-shaped Autoencoder}
The partitioning of the latent features (motivated by Cozzolino et al.'s work ~\cite{cozzolino2018forensictransfer}) and the Y-shaped design of the decoder enables the autoencoder to share valuable information between the classification, segmentation, and reconstruction tasks and thereby improve overall performance by reducing loss. There are three types of loss: activation loss $\mathcal{L}_{act}$, segmentation loss $\mathcal{L}_{seg}$, and reconstruction loss $\mathcal{L}_{rec}$.

Given label $y_i\in \{0,1\}$, activation loss measures the accuracy of partitioning in the latent space on the basis of the activation of the two halves of the encoded features:
\begin{equation}
\mathcal{L}_{act} = \frac{1}{N}\sum_{i}|a_{i,1} - y_i| +  |a_{i,0} - (1 - y_i)|,
\end{equation}
where $N$ is the number of samples, $a_{i,0}$ and $a_{i,1}$ are the activation values and defined as the $L_1$ norms of the corresponding halves of the latent features, $h_{i,0}$ and $h_{i,1}$ (given $K$ is the number of features of $\{h_{i,0}|h_{i,1}\}$):
\begin{equation}
a_{i,c} = \frac{1}{2K}\|h_{i,c}\|_1, ~~ c \in \{0,1\}.
\end{equation}
This ensures that, given an input $x_i$ of class $c$, the corresponding half of the latent features $h_{i,c}$ is activated ($a_{i,c} > 0$). The other half, $h_{i,1-c}$, remains quiesced ($a_{i,1-c} = 0$). To force the two decoders, $D_{seg}$ and $D_{rec}$, to learn the right decoding schemes, we set the off-class part to zero before feeding it to the decoders ($a_{i,1-c} := 0$).

We utilize cross-entropy loss as the segmentation loss to measure the agreement between the segmentation mask ($s_i=\mathcal{D}_{seg}(\{h_{i,0}|h_{i,1}\})$) and the ground-truth mask ($m_i$) corresponding to input $x_i$:
\begin{equation}
\mathcal{L}_{seg} = \frac{1}{N}\sum_{i}\|m_i\log(s_i) + (1 - m_i)\log(1 - s_i)\|_1.
\end{equation}

The reconstruction loss uses the $L_2$ distance to measure the difference between the reconstructed image ($\hat{x}=\mathcal{D}_{rec}(\{h_{i,0}|h_{i,1}\})$) and the original one ($x_i$). For N samples, the reconstruction loss is
\begin{equation}
\mathcal{L}_{rec} = \frac{1}{N}\sum_{i}\|x_i - \hat{x}_i\|_2.
\end{equation}

The total loss is the weighted sum of the three activation losses:
\begin{equation}
\mathcal{L} = \gamma_{act}\mathcal{L}_{act} + \gamma_{seg}\mathcal{L}_{seg} + \gamma_{rec}\mathcal{L}_{rec}.
\end{equation}
Unlike Cozzolino et al.~\cite{cozzolino2018forensictransfer}, we set the three weights equal to each other (equal to 1). This is because the classification task and the segmentation task are equally important, and the reconstruction task plays an important role in the segmentation task. We experimentally compared the effects of the different settings (described below).

\subsection{Implementation}
The Y-shaped autoencoder was implemented as shown in Figure~\ref{figure:network}. It is a fully connected CNN using $3 \times 3$ convolutional windows (for the encoder) and $3 \times 3$ deconvolutional windows (for the decoder) with a stride of 1 interspersed with a stride of 2. Following each convolutional layer is a batch normalization layer~\cite{ioffe2015batch} and a rectified linear unit (ReLU)~\cite{nair2010rectified}. The selection block allows only the true half of the latent features ($h_{i,y_i}$) to pass by and zeros out the other half ($h_{i,1-y_i}$). Therefore, the decoders ($\mathcal{D}_{seg}, \mathcal{D}_{rec}$) are forced to decode only the true half of the latent features. The dimension of the embedding is 128, which has been shown to be optimal~\cite{cozzolino2018forensictransfer}. For the segmentation branch ($\mathcal{D}_{seg}$), a softmax activation function at the end is used to output segmentation maps. For the reconstruction branch ($\mathcal{D}_{rec}$), a hyperbolic tangent function (tanh) is used to shape the output into the range $[-1,1]$. For simplicity, we directly feed normalized images into the autoencoder without converting them into residual images~\cite{cozzolino2018forensictransfer}. Further work will focus on investigating the benefits of using residual images in the classification and segmentation tasks.

Following Cozzolino et al.'s work \cite{cozzolino2018forensictransfer}, we trained the network using the ADAM optimizer~\cite{kingma2014adam} with a learning rate of 0.001, a batch size of 64, betas of 0.9 and 0.999, and epsilon equal to $10^{-8}$.

\begin{table*}[th!]
\centering
\caption{Design of training and testing datasets.}
\label{tab:dataset}
\begin{tabular}{l|llcc}
\multicolumn{1}{c|}{\textbf{Name}} & \multicolumn{1}{c}{\textbf{Source dataset}} & \multicolumn{1}{c}{\textbf{Description}} & \textbf{\begin{tabular}[c]{@{}c@{}}Manipulation\\ Method\end{tabular}} & \textbf{\begin{tabular}[c]{@{}c@{}}Number\\ of Videos\end{tabular}} \\ \hline
Training & \begin{tabular}[c]{@{}l@{}}FaceForensics\\ \textbf{Source-to-Target}\end{tabular} & \begin{tabular}[c]{@{}l@{}}Training set used for\\ all tests\end{tabular} & Face2Face & 704 $\times 2$ \\ \hline
Test 1 & \begin{tabular}[c]{@{}l@{}}FaceForensics\\ \textbf{Source-to-Target}\end{tabular} & \begin{tabular}[c]{@{}l@{}}Match condition for\\ seen attack\end{tabular} & Face2Face & 150 $\times 2$ \\ \hline
Test 2 & \begin{tabular}[c]{@{}l@{}}FaceForensics\\ \textbf{Self-Reenactment}\end{tabular} & \begin{tabular}[c]{@{}l@{}}Mismatch condition for\\ seen attack\end{tabular} & Face2Face & 150 $\times 2$ \\ \hline
Test 3 & \begin{tabular}[c]{@{}l@{}}FaceForensics++\\ \textbf{Deepfakes}\end{tabular} & \begin{tabular}[c]{@{}l@{}}Unseen attack (deep-\\learning-based)\end{tabular} & Deepfakes & 140 $\times 2$ \\ \hline
Test 4 & \begin{tabular}[c]{@{}l@{}}FaceForensics++\\ \textbf{FaceSwap}\end{tabular} & \begin{tabular}[c]{@{}l@{}}Unseen attack (computer-\\graphics-based)\end{tabular} & FaceSwap & 140 $\times 2$
\end{tabular}
\end{table*}

\begin{table*}[th!]
\centering
\caption{Settings for autoencoder.}
\label{tab:design}
\begin{tabular}{c|lccccl}
\textbf{No.} & \multicolumn{1}{c}{\textbf{Method}} & \textbf{Depth} & \textbf{\begin{tabular}[c]{@{}c@{}}Seg.\\ weight\end{tabular}} & \textbf{\begin{tabular}[c]{@{}c@{}}Rec.\\ weight\end{tabular}} & \textbf{\begin{tabular}[c]{@{}c@{}}Rec.\\ loss\end{tabular}} & \multicolumn{1}{c}{\textbf{Comments}} \\ \hline
1 & \textit{FT\_Res} & Shallower & 0.1 & 0.1 & L1 & \begin{tabular}[c]{@{}l@{}}Re-implementation of ForensicsTransfer~\cite{cozzolino2018forensictransfer} using \\ residual images as input\end{tabular} \\ \hline
2 & \textit{FT} & Shallower & 0.1 & 0.1 & L1 & \begin{tabular}[c]{@{}l@{}}Re-implementation of ForensicsTransfer~\cite{cozzolino2018forensictransfer} using \\ normal images as input\end{tabular} \\ \hline
3 & \textit{Deeper\_FT} & Deeper & 0.1 & 0.1 & L1 & \begin{tabular}[c]{@{}l@{}}Proposed deeper version of \textit{FT} (\textit{Proposed\_Old} \\ method without segmentation branch)\end{tabular} \\ \hline
4 & \textit{Proposed\_Old} & Deeper & 0.1 & 0.1 & L1 & \begin{tabular}[c]{@{}l@{}}Proposed method using ForensicsTransfer \\ settings\end{tabular} \\ \hline
5 & \textit{No\_Recon} & Deeper & 1 & 1 & L2 & Proposed method without reconstruction branch \\ \hline
6 & \textbf{\textit{Proposed\_New}} & Deeper & 1 & 1 & L2 & Complete proposed method with new settings
\end{tabular}
\end{table*}

\section{Experiments}
\subsection{Databases}
We evaluated our proposed network using two databases: FaceForensics~\cite{rossler2018faceforensics} and FaceForensics++~\cite{rossler2019faceforensics++}. The FaceForensics database contains 1004 real videos collected from YouTube and their corresponding manipulated versions, which are divided into two sub-datasets:
\begin{itemize}
\item Source-to-Target Reenactment dataset containing 1004 fake videos created using the Face2Face method~\cite{thies2016face2face}; in each input pair for reenactment, the source video (the attacker) and the target video (the victim) are different.
\item Self-Reenactment dataset containing another 1004 fake videos created again using the Face2Face method; in each input pair for reenactment, the source and target videos are the same. Although this dataset is not meaningful from the attacker's perspective, it does present a more challenging benchmark than does the Source-to-Target Reenactment dataset.
\end{itemize}
Each dataset was split into 704 videos for training, 150 for validation, and 150 for testing. The database also provided segmentation masks corresponding to manipulated videos. Three levels of compression based on the H.264 codec\footnote{\url{http://www.h264encoder.com/}} were used: no compression, light compression (quantization = 23), and strong compression (quantization = 40).

The FaceForensics++ database is an enhanced version of the FaceForensics database and includes the Face2Face dataset plus the FaceSwap\footnote{\url{https://github.com/MarekKowalski/FaceSwap/}} dataset (graphics-based manipulation) and the DeepFakes\footnote{\url{https://github.com/deepfakes/faceswap/}} dataset (deep-learning-based manipulation)~\cite{rossler2019faceforensics++}. It contains 1,000 real videos and 3,000 manipulated videos (1,000 in each dataset). Each dataset was split into 720 videos for training, 140 for validation, and 140 for testing. The same three levels of compression based on the H.264 codec were used with the same quantization values.

For simplicity, we used only videos with light compression (quantization = 23). Images were extracted from videos using Cozzolino et al.'s settings~\cite{cozzolino2018forensictransfer}: 200 frames of each training video were used for training, and 10 frames of each validation and testing video were used for validation and testing, respectively. There is no detailed description of the rules for frame selection, so we selected the first (200 or 10) frames of each video and cropped the facial areas. For all databases, we applied normalization with $\text{mean} = (0.485, 0.456, 0.406)$ and $\text{standard deviation} = (0.229, 0.224, 0.225)$; these values have been widely used in the ImageNet Large Scale Visual Recognition Challenge~\cite{ILSVRC15}. We did not apply any data augmentation to the trained datasets.

The training and testing datasets were designed as shown in Table~\ref{tab:dataset}. For the Training, Test 1, and Test 2 datasets, the Face2Face method~\cite{rossler2018faceforensics} was used to create manipulated videos. Images in Test 2 were harder to detect than those in Test 1 since the source and target videos used for reenactment were the same, meaning that the reenacted video frames had better quality. Therefore, we call Test 1 and Test 2 the \textbf{match} and \textbf{mismatch} conditions for a seen attack. Test 3 used the Deepfake attack method while Test 4 used the FaceSwap attack method, presented in the FaceForensics++ database~\cite{rossler2019faceforensics++}. These both attack methods were not used to create the training set, therefore they were considered as unseen attacks. For the classification task, we calculated the accuracy and equal error rate (EER) of each method. For the segmentation task, we used pixel-wise accuracy between ground-truth masks and segmentation masks. The \textit{FT\_Res}, \textit{FT}, and \textit{Deeper\_FT} method could not perform the segmentation task. All the results were at the image level.

\subsection{Training Y-shaped Autoencoder}
To evaluate the contributions of each component in the Y-shaped autoencoder, we designed the settings as shown in Table~\ref{tab:design}. The \textit{FT\_Res} and \textit{FT} methods are re-implementations of Cozzolino et al.'s method with and without using residual images~\cite{cozzolino2018forensictransfer}. They can also be understood as the Y-shaped autoencoder without the segmentation branch. The \textit{Deeper\_FT} method is a deeper version of \textit{FT}, which has the same depth as the proposed method. The \textit{Proposed\_Old} method is the proposed method using weighting settings from Cozzolino et al.'s work~\cite{cozzolino2018forensictransfer}, the \textit{No\_Recon} method is the version of the proposed method without the reconstruction branch, and the \textit{Proposed\_New} method is the complete proposed method with the Y-shaped autoencoder using equal losses for the three tasks and the mean squared error for reconstruction loss.

Since shallower networks take longer to converge than deeper ones, we trained the shallower ones with 100 epochs and the deeper ones with 50 epochs. For each method, the training stage with the highest accuracy for the classification task and a reasonable segmentation loss (if available) was used to perform all the tests described in this section.

\subsection{Dealing with Seen Attacks}
The results for the match and mismatch conditions for seen attacks are respectively shown in Tables~\ref{tab:test_1} (Test 1) and ~\ref{tab:test_2} (Test 2). The deeper networks (the last four) had substantially better classification performance than the shallower ones (the first two) proposed by Cozzolino et al. ~\cite{cozzolino2018forensictransfer}. Among the four deeper networks, there were no substantial differences in their performances on the classification task. For the segmentation task, the \textit{No\_Recon} and \textit{Proposed\_New} methods, which used the new weighting settings, had higher accuracy than the \textit{Proposed\_Old} method, which used the old weighting settings. 

\begin{table}[th!]
\centering
\caption{Results for Test 1 (image level).}
\label{tab:test_1}
\begin{tabular}{l|cc|c}
\multicolumn{1}{c|}{\multirow{2}{*}{\textbf{Method}}} & \multicolumn{2}{c|}{\textbf{Classification}} & \textbf{Segmentation} \\ \cline{2-4} 
\multicolumn{1}{c|}{} & \textbf{Acc (\%)} & \textbf{EER (\%)} & \textbf{Acc (\%)} \\ \hline
\textit{FT\_Res} & 82.30 & 14.53 & \_ \\ \hline
\textit{FT} & 88.43 & 11.60 & \_ \\ \hline
\textit{Deeper\_FT} & \textbf{93.63} & 7.20 & \_ \\ \hline
\textit{Proposed\_Old} & 92.60 & 7.40 & 81.40 \\ \hline
\textit{No\_Recon} & 93.40 & \textbf{7.07} & 89.21 \\ \hline
\textbf{\textit{Proposed\_New}} & 92.77 & 8.18 & \textbf{90.27}
\end{tabular}
\end{table}

\begin{table}[th!]
\centering
\caption{Results for Test 2 (image level).}
\label{tab:test_2}
\begin{tabular}{l|cc|c}
\multicolumn{1}{c|}{\multirow{2}{*}{\textbf{Method}}} & \multicolumn{2}{c|}{\textbf{Classification}} & \textbf{Segmentation} \\ \cline{2-4} 
\multicolumn{1}{c|}{} & \textbf{Acc (\%)} & \textbf{EER (\%)} & \textbf{Acc (\%)} \\ \hline
\textit{FT\_Res} & 82.33 & 15.07 & \_ \\ \hline
\textit{FT} & 87.33 & 12.03 & \_ \\ \hline
\textit{Deeper\_FT} & 92.70 & \textbf{7.80} & \_ \\ \hline
\textit{Proposed\_Old} & 91.83 & 8.53 & 81.40 \\ \hline
\textit{No\_Recon} & \textbf{92.83} & 8.29 & 89.10 \\ \hline
\textbf{\textit{Proposed\_New}} & 92.50 & 8.07 & \textbf{90.20}
\end{tabular}
\end{table}

The performances of all methods was slightly degraded when dealing with the mismatch condition for seen attacks. The \textit{FT\_Res} and \textit{Proposed\_New} methods had the best adaptation ability, as indicated by the lower degradation in their scores. This indicates the importance of using residual images (for the \textit{FT\_Res} method) and of using the reconstruction branch (for the Y-shaped autoencoder with new weighting settings: \textit{Proposed\_New} method). The reconstruction branch also helped the \textit{Proposed\_New} method achieve the highest score on the segmentation task. 

\subsection{Dealing with Unseen Attacks}
\subsubsection{Evaluation using pre-trained model}
When encountering unseen attacks, all six methods had substantially lower accuracies and higher EERs, a shown in Tables~\ref{tab:test_3} (Test 3) and~\ref{tab:test_4} (Test 4). In Test 3, the shallower methods had better adaptation ability, especially the \textit{FT\_Res} method, which uses residual images. The deeper methods, which had a greater chance of being over-fitted, had nearly random classification results. In Test 4, although all methods suffered from nearly random classification accuracies, their better EERs indicated that the decision thresholds had been moved.

A particularly interesting finding was in the segmentation results. Although degraded, the segmentation accuracies were still high, especially in Test 4, in which FaceSwap copied the facial area from the source faces to the target ones using a computer-graphics method. When dealing with unseen attacks, this segmentation information could thus be an important clue in addition to the classification results for judging the authenticity of the queried images and videos.

\begin{table}[th!]
\centering
\caption{Results for Test 3 (image level).}
\label{tab:test_3}
\begin{tabular}{l|cc|c}
\multicolumn{1}{c|}{\multirow{2}{*}{\textbf{Method}}} & \multicolumn{2}{c|}{\textbf{Classification}} & \textbf{Segmentation} \\ \cline{2-4} 
\multicolumn{1}{c|}{} & \textbf{Acc (\%)} & \textbf{EER (\%)} & \textbf{Acc (\%)} \\ \hline
\textit{FT\_Res} & \textbf{64.75} & \textbf{30.71} & \_ \\ \hline
\textit{FT} & 62.61 & 37.43 & \_ \\ \hline
\textit{Deeper\_FT} & 51.21 & 42.71 & \_ \\ \hline
\textit{Proposed\_Old} & 53.75 & 42.00 & 70.18 \\ \hline
\textit{No\_Recon} & 51.96 & 42.45 & \textbf{70.43} \\ \hline
\textbf{\textit{Proposed\_New}} & 52.32 & 42.24 & 70.37
\end{tabular}
\end{table}

\begin{table}[th!]
\centering
\caption{Results for Test 4 without fine-tuning (image level).}
\label{tab:test_4}
\begin{tabular}{l|cc|c}
\multicolumn{1}{c|}{\multirow{2}{*}{\textbf{Method}}} & \multicolumn{2}{c|}{\textbf{Classification}} & \textbf{Segmentation} \\ \cline{2-4} 
\multicolumn{1}{c|}{} & \textbf{Acc (\%)} & \textbf{EER (\%)} & \textbf{Acc (\%)} \\ \hline
\textit{FT\_Res} & 53.50 & 43.10 & \_ \\ \hline
\textit{FT} & 52.29 & 41.79 & \_ \\ \hline
\textit{Deeper\_FT} & 53.39 & 37.00 & \_ \\ \hline
\textit{Proposed\_Old} & \textbf{56.82} & 36.29 & 84.23 \\ \hline
\textit{No\_Recon} & 54.86 & 35.86 & \textbf{84.86} \\ \hline
\textbf{\textit{Proposed\_New}} & 54.07 & \textbf{34.04} & 84.67
\end{tabular}
\end{table}

\subsubsection{Fine-tuning using small amount of data}
We used the validation set (a small set normally used for selecting hyper-parameters in training that differs from the test set) of the FaceForensics++ - FaceSwap dataset~\cite{rossler2019faceforensics++} for fine-tuning all the methods. To ensure that the amount of data was small, we used only ten frames for each video. We divided the dataset into two parts: 100 videos of each class for training and 40 of each class for evaluation. We trained them using 50 epochs and selected the best models on the basis of their performance on the evaluation set.

\begin{table}[th]
\centering
\caption{Results for Test 4 after fine-tuning (image level).}
\label{tab:test_4_finetuning}
\begin{adjustbox}{width=0.48\textwidth}
\begin{tabular}{l|cc|c}
\multicolumn{1}{c|}{\multirow{2}{*}{\textbf{Method}}} & \multicolumn{2}{c|}{\textbf{Classification}} & \textbf{Segmentation} \\ \cline{2-4} 
\multicolumn{1}{c|}{} & \textbf{Acc (\%)} & \textbf{EER (\%)} & \textbf{Acc (\%)} \\ \hline
\textit{FT\_Res} & 80.04 ($\uparrow$ 26.54) & 17.57 \textbf{($\downarrow$ 25.53)} & \_ \\ \hline
\textit{FT} & 70.89 ($\uparrow$ 18.60) & 25.56 ($\downarrow$ 16.23) & \_ \\ \hline
\textit{Deeper\_FT} & 82.00 ($\uparrow$ 28.61) & 17.33 ($\downarrow$ 19.67) & \_ \\ \hline
\textit{Proposed\_Old} & 78.57 ($\uparrow$ 21.75) & 20.79 ($\downarrow$ 15.50) & 84.39 ($\uparrow$ 0.16)\\ \hline
\textit{No\_Recon} & 82.93 ($\uparrow$ 28.07) & 16.93 ($\downarrow$ 18.93) & 92.60 ($\uparrow$ 7.74) \\ \hline
\textbf{\textit{Proposed\_New}} & \textbf{83.71 ($\uparrow$ 29.64)} & \textbf{15.07} ($\downarrow$ 18.97) & \textbf{93.01 ($\uparrow$ 8.34)}
\end{tabular}
\end{adjustbox}
\end{table}

The results after fine-tuning for Test 4 are shown in Table~\ref{tab:test_4_finetuning}. Their classification and segmentation accuracies increased around 25\% and 8\%, respectively, which are remarkable compared with the small amount of data used. The one exception was the \textit{Proposed\_Old} method -- its segmentation accuracy did not improve. The \textit{FT\_Res} method had better adaptation than the \textit{FT} one, which supports Cozzolino et al.'s claim~\cite{cozzolino2018forensictransfer}. The \textit{Proposed\_New} method had the highest transferability against unseen attacks as evidenced by the results in Table~\ref{tab:test_4_finetuning}.

\section{Conclusion}
The proposed convolutional neural network with a Y-shaped autoencoder demonstrated its effectiveness for both classification and segmentation tasks without using a sliding window, as is commonly used by classifiers. Information sharing among the classification, segmentation, and reconstruction tasks improved the network's overall performance, especially for the mismatch condition for seen attacks. Moreover, the autoencoder can quickly adapt to deal with unseen attacks by using only a few samples for fine-tuning. Future work will mainly focus on investigating the effect of using residual images~\cite{cozzolino2018forensictransfer} on the autoencoder's performance, processing high-resolution images without resizing, improving its ability to deal with unseen attacks, and extending it to the audiovisual domain.

\section*{Acknowledgement}
This research was supported by JSPS KAKENHI Grant Number JP16H06302, JP18H04120, and JST CREST Grant Number JPMJCR18A6, Japan.

{\small
\bibliographystyle{ieee}
\bibliography{main}
}

\end{document}